# Extracting real estate values of rental apartment floor plans using graph convolutional networks

Atsushi Takizawa

Osaka Metropolitan University

**Keywords**

Rental apartment, rent, hedonic model, floor plan, access graph, graph convolutional networks

**Abstract**

Access graphs that indicate adjacency relationships from the perspective of flow lines of rooms are extracted automatically from a large number of floor plan images of a family-oriented rental apartment complex in Osaka Prefecture, Japan, based on a recently proposed access graph extraction method with slight modifications. We define and implement a graph convolutional network (GCN) for access graphs and propose a model to estimate the real estate value of access graphs as the floor plan value. The model, which includes the floor plan value and hedonic method using other general explanatory variables, is used to estimate rents, and their estimation accuracies are compared. In addition, the features of the floor plan that explain the rent are analyzed from the learned convolution network. Therefore, a new model for comprehensively estimating the value of real estate floor plans is proposed and validated. The results show that the proposed method significantly improves the accuracy of rent estimation compared to that of conventional models, and it is possible to understand the specific spatial configuration rules that influence the value of a floor plan by analyzing the learned GCN.

# Introduction

Attempts have been made to evaluate housing preferences and satisfaction statistically and quantitatively, although they are subjective. Indices such as neighborhood, location, size, and housing type have been used frequently (Adir et al. 1996, Jim and Chen 2009, Lovejoy et al. 2010, Michelson 1977, Valente et al. 2005). Some studies evaluated houses based on more detailed attributes such as the number of rooms, size, shape, room size, and facilities (Hofman et al. 2006, Goodman et al. 2007). The latter study hypothesized that housing preferences are influenced by a combination of many detailed factors such as design, usability, and spatial configuration, and this may be true despite individual differences. However, quantifying these factors is difficult when they are considered in detail. For example, Jirovec et al. (1984) applied the building evaluation index proposed by Marans and Spreckelmeyer (1982) to evaluate housing for the elderly and attempted to explain it using a hedonic model (Rosen 1974) that includes aesthetic attributes. However, they pointed out that the explanatory power of the location variables was strong and that of the aesthetic ones was weak. Hofman et al. (2006) and Goodman et al. (2007) evaluated the number of rooms, area, and defined shape, which are relatively easy to quantify, as independent variables. It is difficult to quantify space at a level of complexity that can be perceived by humans, as is the built environment, which includes housing.

Many studies focused on quantifying these complexities of space. Among them, Space Syntax (Hillier and Hanson 1984) was probably the most successful. Several analysis methods are used in Space Syntax, but for floor plan analysis, the space is divided into appropriate units (i.e., nodes) and access graphs are constructed by connecting rooms and spaces connected in terms of flow lines by edges for extracting the indicators of graph depth and centrality. The features of residential floor plans have been extracted for various types of housing, which include traditional houses (Hanson 1998), high-rise apartment buildings (Hanazato et al. 2005), and detached houses of famous architects (Ostwald and Dawes 2018).

Thus far, several technologies were developed to overcome the abovementioned problems. The LIFULL HOME'S dataset (LIFULL Co., Ltd. 2015) is a large dataset of Japanese real estate, and after it was released, several methods were developed to extract access graphs from images automatically using deep learning (Yamada et al. 2021, Yamasaki et al. 2018) and other similar techniques. Besides these methods, deep learning methods were developed, one of which is a type of deep learning method that convolves graphs and images (Niepert et al. 2016).

Given this background, this study uses the LIFULL HOME'S dataset to extract access graphs automatically from a large number of floor plan images of family-oriented rental housing in Osaka Prefecture using a slightly modified version of the method described by Yamada et al. (2021). Then, we define and implement a graph convolution network (GCN) for the access graph and propose a model that estimates the real estate value of a homomorphic access graph as the floor plan value (FPV) after learning rent estimation. The model with FPV and the hedonic method that uses other general explanatory variables are used for estimating rents, and their estimation accuracies are compared. We also analyzed the features of the floor plans that explain the rents from the learned convolution network. Therefore, we proposed and validated a new model for comprehensively estimating the value of real estate floor plans. Takizawa et al. (2008) selected a family-type rental housing as the subject of the analysis for the following reasons: (1) Ease of obtaining a large amount of data; (2) the spatial configuration has a moderate complexity and diversity but does not have the same degree of freedom as detached houses, and there is less extrapolated data for the estimation model; and (3) the supply of multifamily housing has been increasing in Japan since the 1980s (Value Management Institute 2013) and has a large social and economic impact.

Next, we summarize the related research. There have been many studies on estimating the value of real estate using hedonic methods. With the development of deep learning in the late 2010s, the interior and exterior of buildings have been used as image data (Ahmed et al. 2016, You et al. 2017, Glaeser et al. 2018, Poursaeed et al. 2018) and floor plans (Solovev and Pröllochs 2021, Hattori et al. 2021) in hedonic models. Such studies are the common motivation for this study. The approach that uses floor plan images directly does not suffer from the risk of access graph misclassification. For models trained using floor plan images, decision making in the learned model is represented by image feature points such as speeded up robust features (SURFs) (Bay et al. 2008), which are not interpretable from an architectural or real estate perspective; the author believes that this type of model is a black box that improves prediction accuracy.

The evaluation of housing preferences and ease of use is not only estimated from hedonic methods but also via the subjective evaluation of floor plan images. Gao et al. (2013) and Tamura and Fang (2022) used Space Syntax to explain the evaluation results, and they revealed that the degree of privacy affects the evaluation of certain resident groups. Kato et al. (2020) and Narahara and Yamasaki (2022) conducted subjective evaluations based on the aforementioned floor plan images of the LIFULL HOME'S dataset and they combined multiple deep learning methods in their evaluation prediction models. Narahara and Yamasaki used a large sample of over 3,000 subjects and inferred that the prediction model has a certain degree of reliability. The number of presented floor plans was limited because of the nature of the questionnaire, and the relationship with the rent was not evaluated.

GCNs have been applied since the 2010s in fields such as chemistry (Duvenaud et al. 2015). Based on the aforementioned literature review and to the best of the author's knowledge, this is the first study that constructed and analyzed rent estimation models using GCNs based on the graphs of floor plans.

The framework of this study is illustrated in Figure 1.

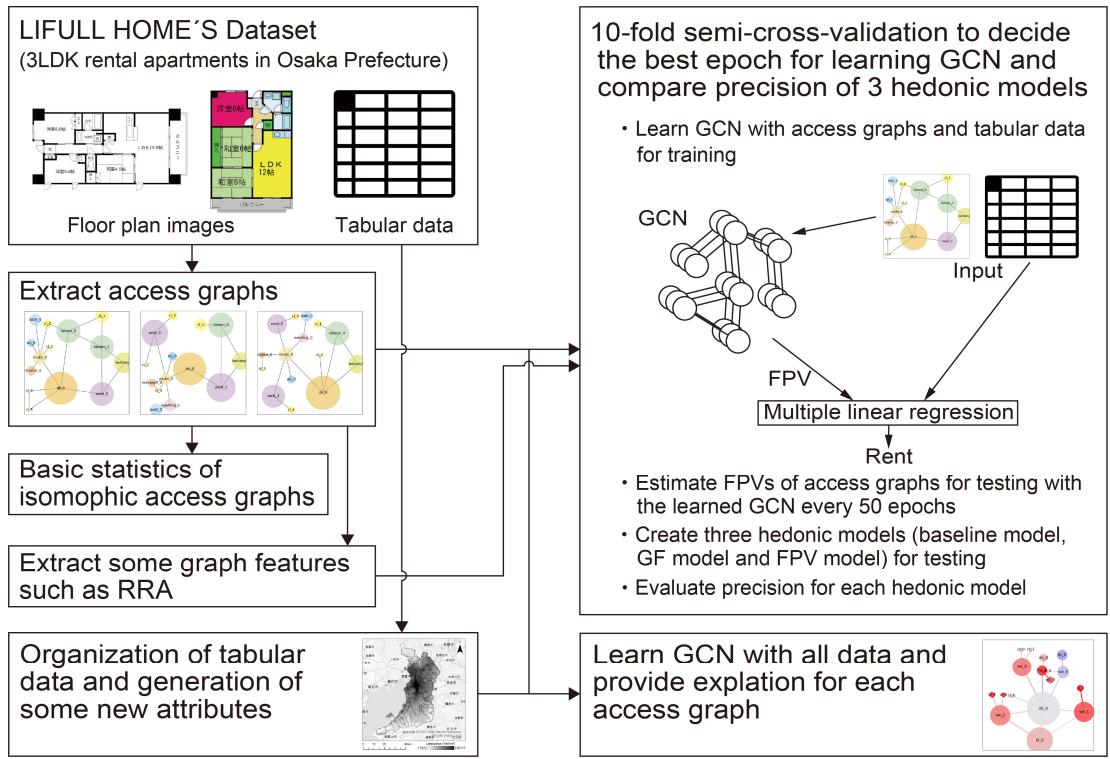

Figure 1. Framework of this study.

## Dataset preparation

This section describes the procedure and results of extracting the floor plan access graphs.

**Step 1: Define an access graph**

In this study, the floor plan of a house is represented as an access graph that indicates adjacency relationships from the perspective of room flow lines. A GCN is used to learn FPVs in accordance with the access graph extraction method developed by Yamada et al. (2021). Figure 2 shows an example of a floor plan and its access graph. The access graph is represented as a graph with the node labels of room types; 10 labels are used in this study: balcony (bl), bathroom (bt), closet (cl), living room, dining area, and kitchen (dk), entrance (en), hallway (hw), lavatory (la),

Japanese room (ja), toilet (to), and Western room (we). The notation "_0,_1,…" after the label indicates a sequential number that distinguishes the nodes with the same label. There are two types of room connections: (1) the rooms are separated by a door or window as a boundary, and (2) the rooms are recognized as different spaces without a physical partition, such as an entrance and a hallway; however, they are not distinguished in the access graph. Openings are treated as walls because they are not connected to other spaces by edges even if they are recognized, except for the part that serves as a border with a balcony.

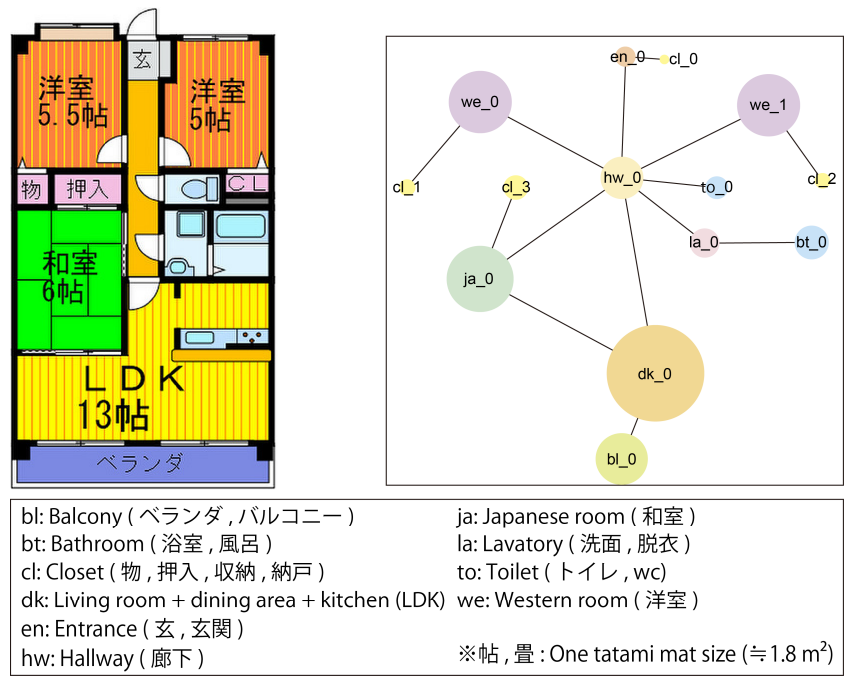

Figure 2. Examples of a floor plan and its access graphs.

## Step 2: Filtering data and extracting access graphs

This study used the LIFULL HOME'S dataset described above, and this dataset includes three types of data. Among them, this study uses the snapshot data of rentals that include various pieces of information on approximately 5.33 million rental properties nationwide as of September 2015 and the high-resolution floor plan image data of those properties.

Table 1 lists the procedure for extracting the data, which includes the access graphs necessary for this study. In Procedure 2, we use the adjacency graph extraction method developed in the literature (Yamada et al. 2021) with some modifications. The improvements and accuracy evaluations are presented in Appendix A1. The graph isomorphism test (Alfred et al. 1974) in Procedure 3 determines whether the two graphs have the same topology and selects only unique access graphs. In this study, the isomorphism test considers the type of node (usage). The isomorphic function of NetworkX (NetworkX developers 2014), which is a Python graph library, is used by setting its node-matching option to the list of node labels. The isomorphism test is performed by setting the node-matching option to the list of node labels. Consequently, 7970 different access graphs are extracted. A total of 19,998 access graphs are aggregated using these isomorphic graphs. A histogram that shows the number of corresponding properties in the descending order is shown in Figure 3. The access graphs are concentrated in certain types. More than half of the access graphs have one property and they tend to have a small number of properties. Access graphs that are easily misjudged tend to have a small number of properties, and the actual number of unique access graph types is likely to be less than 7980. However, the overall accuracy is approximately 80% (see Appendix A1), and the analysis is conducted by including the case where the number of properties is one. Step 3 explains the eight attributes considered in Procedure 4.

Table 1. Procedures for preparing the data for analysis.

| Procedure | Description | Number of applicable properties | Number of unique access graphs |
|---|---|---|---|
| 1 | Select properties from the LIFULL HOME'S dataset based on the following criteria: "Prefecture = Osaka Prefecture," "Property type = Residential rental condominium or apartment," "Room type = 3LDK," and high-precision floor plan images exist. | 46,466 | - |
| 2 | Extract access graphs from floor plan images. | 19,998 | - |
| 3 | Isomorphism test of access graphs. | ↑ | 7970 |
| 4 | Among the extracted access graphs, select properties that have all eight attributes presented in the tabular data and delete one property with a large outlier in rent. | 15,323 | 7865 |

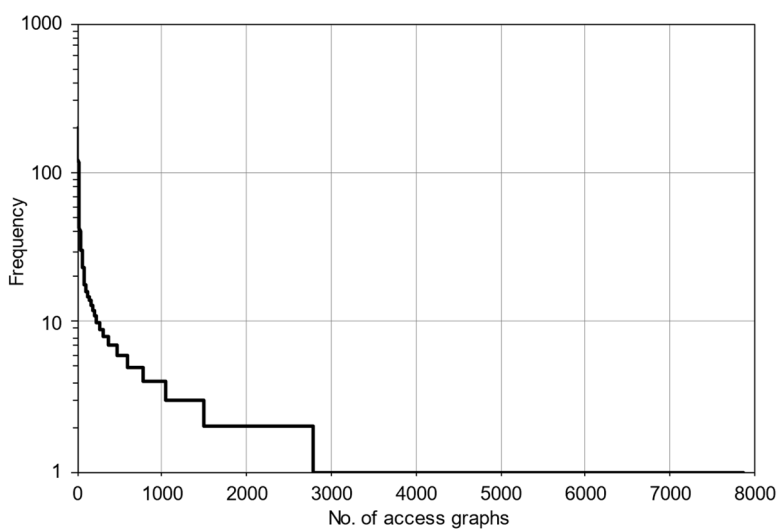

Figure 3. Number of properties in each isomorphic access graph sorted in the descending order of frequency.

Figure 4 shows floor plans with the top three most frequently accessed graphs. Many recent floor plans in Japan do not have a Japanese room; however, top access graphs are those with a Japanese room. Table 2 indicates that the average year of building completion in the data is 1994, which implies that many of them are approximately 30 years old. The two types of room layouts (Top 2 and Top 3) are similar; however, they differ in terms of whether they have a storage space in the hallway.

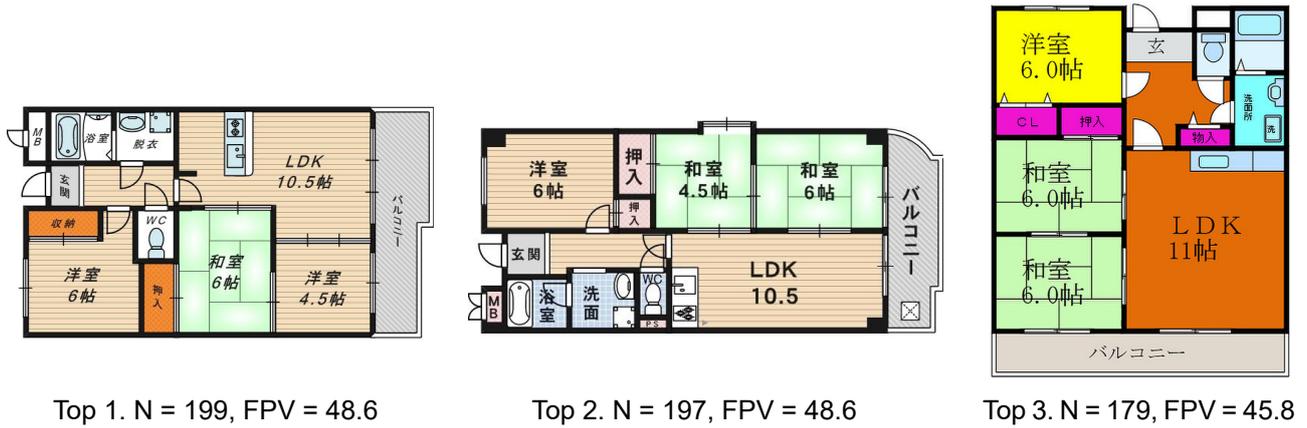

Figure 4. Floor plans with the top three most frequent access graphs.

## Step 3: Prepare tabular data

We organized the common tabular data for using them to train the access graphs and in the final hedonic model. The dataset used contained 71 attributes. Among these, the monthly rent without common expenses was used as the objective variable. As explanatory variables, we used the following items with high coverage: average land price, year of construction, number of passengers per day at the nearest station in 2016 (MLIT, 2016), distance to the nearest station/bus stop, building structure, housing unit area, number of building floors, and number of housing unit floors.

The average land prices were obtained using the following procedure: Public land price data (MLIT 2015a) and prefectural land price survey data (MLIT 2015b) for Osaka and the adjacent prefectures in 2015 were plotted using ArcGIS Pro 2.8. Kriging was performed on these points to fine tune the discontinuous values, and a continuous map of land prices was constructed (Figure 5). We used the zip code boundary polygon data of Osaka Prefecture (Kogyo 2020) to obtain the average land value of the raster data in each polygon because the spatial resolution of each property was limited to the zip code level. Then, this was used as the land value of the corresponding property. Figure 6 shows a map of the 15,323 properties obtained in Step 2 aggregated by each zip code area. Tables 2 and 3 summarize the tabular data used for GCN learning and rent estimation after adding the attributes as indicated.

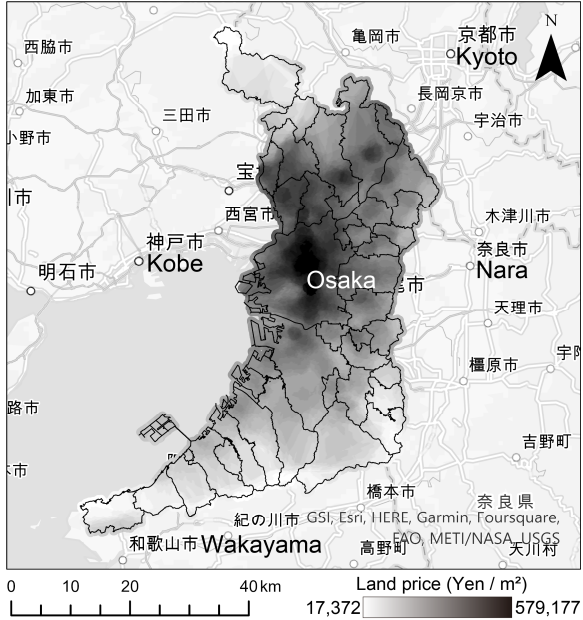 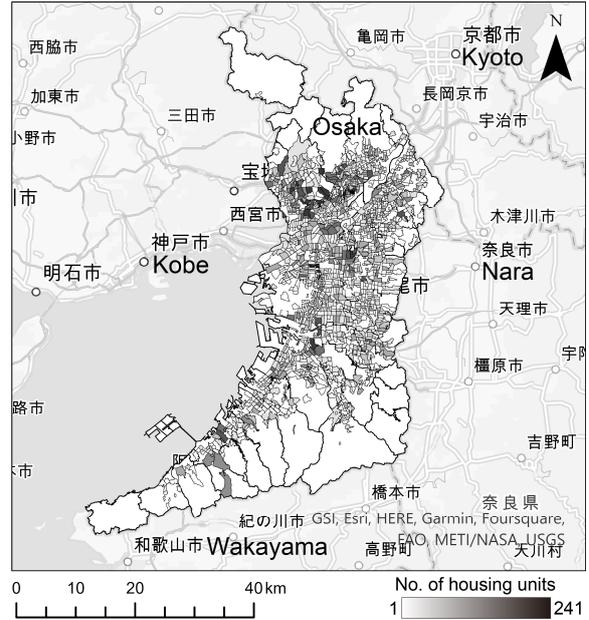

Figure 5. Estimated distribution of land prices in Osaka Prefecture in 2015.

Figure 6. Aggregate number of housing units by zip code area.

Table 2. Basic statistics of quantitative variables (N = 15,323).

| Variable | Description | Min | Mean | Median | Max | SD |
| --- | --- | --- | --- | --- | --- | --- |
| *Area* | Area of housing unit ($m^2$) | 37 | 65.2 | 65 | 159 | 7.3 |
| *Distance* | Distance to nearest station/bus stop (m) | 10 | 865 | 800 | 37,600 | 629 |
| *F_Building* | Number of building floors | 0 | 5.9 | 5 | 50 | 4.0 |
| *F_Dwelling* | Number of floors of housing unit | 1 | 3.6 | 3 | 77 | 2.8 |
| *Land price* | Average land price in zip code area (Yen/$m^2$) | 0 | 187,824 | 184,500 | 557,725 | 75,396 |
| *Passenger* | Number of passengers at nearest station | 0 | 28,447 | 21,604 | 431,007 | 26,638 |
| *Year* | Year of building completion | 1933 | 1994 | 1994 | 2015 | 6.8 |
| *Rent* | Rent without common expenses (Yen/month) | 38,000 | 82,973 | 77,000 | 450,000 | 24,659 |

SD represents standard deviation.

Table 3. Frequency of each type of structure (N = 15,323).

| Structure | Frequency |
| --- | --- |
| Reinforced concrete (RC) | 11,327 |
| Steel construction (S) | 1943 |
| Steel reinforced concrete (SRC) | 1055 |
| Lightweight steel construction (LS) | 672 |
| Autoclaved lightweight aerated concrete (ALC) | 175 |
| Wooden (W) | 124 |
| Precast concrete (PC) | 14 |
| Other (O) | 10 |
| Hard precast concrete (HPC) | 3 |

**Step 4: Extract explicit features of the access graph**

We build a rent estimation model from the access graphs obtained using the GCN; however, for comparison, we also build a rent estimation model with the explicitly given graph features. We call a model with graph features a GF model. The GF model uses the total number of nodes and edges that constitute each access graph (*Num_node and Num_edge*), number of nodes for each room type (*Num_xx*), maximum depth from the entrance (*Depth*), real relative asymmetry (*RRA_xx*), and *H\** as the explanatory variables (Zako 2006) (see Appendix A4). The latter three features were used in the Space Syntax analysis. The maximum depth from the entrance indicates the degree of privacy from outside. *RRA* indicates the degree of isolation of each node in the entire graph, with larger values indicating greater isolation. *H\** indicates the diversity of the RRA distribution for each node in the graph, with larger values indicating more diversity in the RRA for each node, that is, a deeper space. The larger the value, the more diverse is the RRA of each node, that is, the deeper is the space.

In this study, the RRA was obtained for each node as a starting point, except for the closets. However, the node with the highest value was used when there were multiple rooms of the same type. The value was set to zero if there were no corresponding nodes. Owing to the large number of prepared variables, only the variables employed in the GF model after variable selection are listed in Table 4.

Table 4. Basic statistics for graph features used in GF model (N = 15,323).

| Variable | Description | Min | Mean | Median | Max | SD |
| --- | --- | --- | --- | --- | --- | --- |
| *Num_edge* | Total number of edges | 8 | 14.76 | 15 | 25 | 1.884 |
| *Num_bl* | Number of balconies | 0 | 1.243 | 1 | 3 | 0.522 |
| *Num_cl* | Number of closets | 0 | 3.795 | 4 | 9 | 1.325 |
| *Depth* | Max depth from entrance | 3 | 3.889 | 4 | 8 | 0.480 |
| *RRA_bt* | RRA of bath room | 0.678 | 1.428 | 1.430 | 2.345 | 0.155 |
| *RRA_jp* | Maximum RA of Japanese room | 0.000 | 0.859 | 0.941 | 2.278 | 0.294 |
| *RRA_to* | RRA of toilet | 0.677 | 1.061 | 1.001 | 2.345 | 0.207 |
| *H\** | Relative ungelatinized difference factor | 0.278 | 0.700 | 0.701 | 0.888 | 0.061 |

# Extracting floor plan values using the graph convolutional network

This section describes the method to extract FPVs using GCN.

**Step 1: Definition of GCN**

A GCN is used to estimate the real estate value of a floor plan by inputting the floor plan data represented as an access graph. To train this network, the rent of each property should be used as an objective variable; however, the rent is not determined solely from the floor plan, and therefore, the aforementioned table data are also used. Figure 7 shows the learning model that combines a GCN and the multiple regression model. PyG (PyG 2022) is used to implement this model. There are several standard architectures for GCNs in convolutional layers. However, there is no typical architecture for their combination such as VGG (Simonyan and Zisserman 2015) in CNNs for image processing. In this study, we employed a graph network (Zhao et al. 2018) that mimics the structure of VGG, which has a simpler structure; this GCN model performed comparatively well on many graph datasets. The main convolutional layer, ResGatedGraphConv (Bresson and Laurent 2017), which performed well in the preliminary experiments, was also used. However, the first layer is only a more basic GCNConv (Kipf and Welling 2016) because the calculation of the integrated gradient (IG; Sundararajan et al. 2017) is the method used to explain the model described below, and it requires an edge attribute in the input layer for masking. However, ResGatedGraphConv does not allow for an edge attribute.

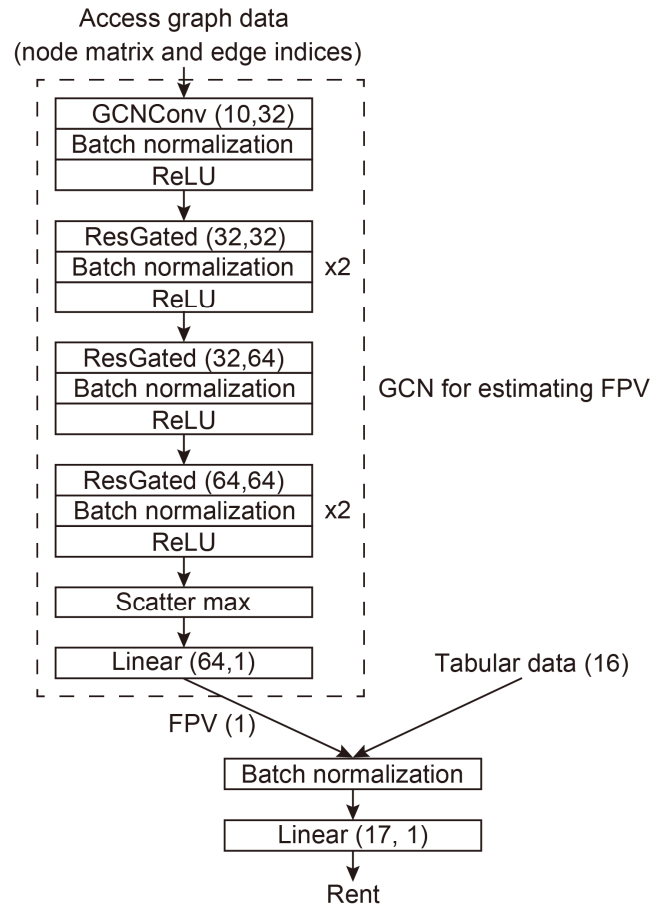

Figure 7. Multiple regression model for training including GCN. Numbers in () are dimensions of (input, output) data.

**Step 2: Training of GCN with all data**

The GCN was performed using the neural network, as shown in Figure 8. The mean squared error (MSE) loss was used as the loss function, and the training settings were optimizer = Adam, learning rate = 0.01, batch size = 1024, and epoch size = 2000. Figure 8 shows the results of the convergence evaluation on the training data using all 15,323 data points. The root mean squared error (RMSE) of the training data at the end of 2000 epochs is 6657 Yen, which is approximately 8% relative to the average rent. The convergence of the model was judged to be satisfactory, and the parameters of the model were stored at every 50 epochs and the real estate value of each floor plan was estimated later.

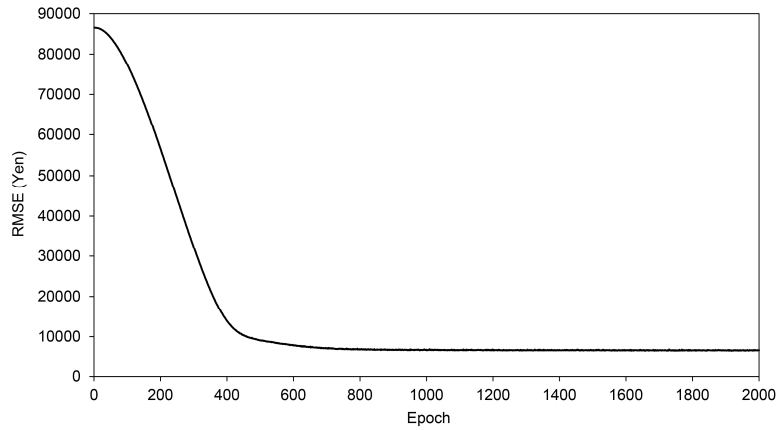

Figure 8. Convergence of errors in the model trained with all data.

**Step 3: Determine the optimal epochs for GCN training using 10-fold semi-cross-validation**

Next, 10-fold semi-cross-validation learning was performed to evaluate the prediction accuracy of the FPV estimation model for each of the 50 epochs. The model for the epoch with the highest prediction accuracy was used to estimate the FPV, and the prediction accuracy was verified. The 15,323 data points were divided randomly and equally into 10 segments, and each segment was divided into training and test data. Next, we trained the model shown in Figure 7 on the training data for each partition using the training parameters, and we saved the model for each of the 50 epochs. Only the GCN portion was extracted from each saved model to estimate the FPV of the test data and a multiple regression model consisting of that value, other attributes, and a constant term created to determine the accuracy (i.e., RMSE). Unlike usual cross-validation, a multiple regression model is trained and validated with only FPV as the unknown data during accuracy validation, which we call semi-cross-validation.

The distribution of the RMSE obtained after repeating this process 10 times is demonstrated in Figure 9. The model with the smallest average RMSE was the FPV trained at the 1550 epoch. We evaluate the accuracy of the baseline and graph attribute models using the same test data.

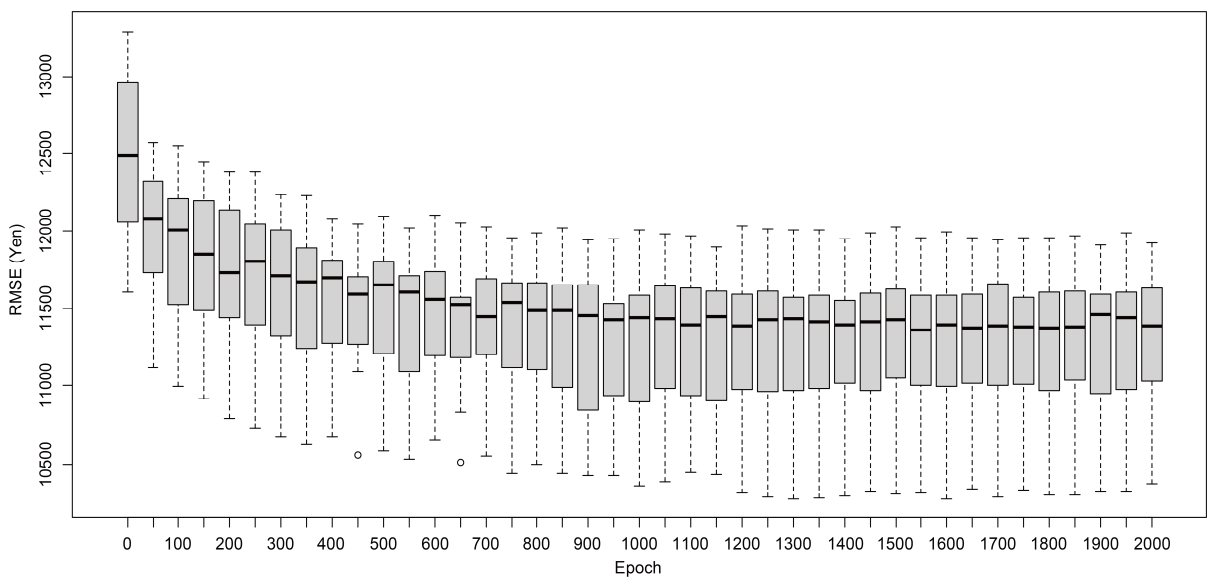

Figure 9. Distribution of RMSEs for 10-fold semi-cross-validations of the rent estimation model built on test data incorporating the FPVs for each epoch.

**Step 4: Distribution of FPVs**

The original values were converted to deviation values for facilitating their positioning because the FPV is a relative value and the absolute value is meaningless. The deviation value is obtained by converting the original data such that the mean is 50 and the standard deviation is 10, which is routinely used in Japan to compare test scores. Let $x, \bar{x}, and\ \sigma$ be the values of the FPVs before conversion, their mean, and standard deviation, respectively. Then, the deviation $FPV(x)$ of a given FPV $x$ is defined by

$$FPV(x) = 10\frac{x - \bar{x}}{\sigma} + 50$$

Figure 10 shows a histogram of the deviation of the FPVs for the 7865 unique access graphs estimated by the model at 1550 epochs when all graphs were trained with the data. The minimum, maximum, and median values were 2.2, 226.9, and 48.6, respectively. The distribution was not normal; however, it was asymmetric with the maximum value having a base longer than the mean value. Hereafter, this deviation in the FPV is referred to as the FPV.

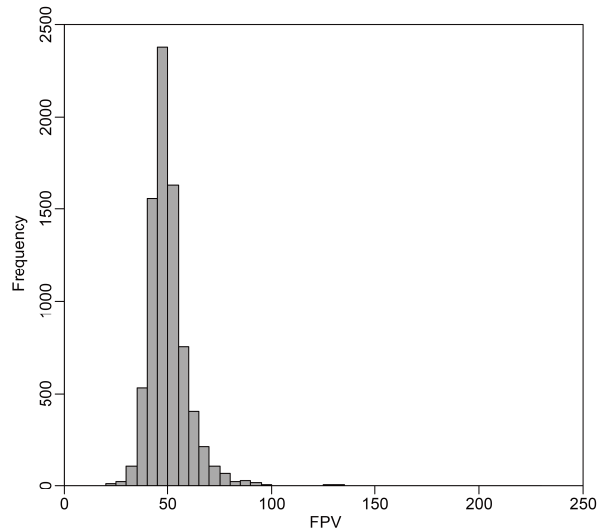

Figure 10. Histogram of deviation in FPVs for the 7865 unique floor plans estimated in the 1550 epoch model.

## Comparison of the accuracies of the three hedonic models

To verify the validity of the hedonic model that considers the FPV model, a rent estimation model that uses only general attributes as the baseline model and a model that adds graph attributes to the baseline model (GF model) were compared in terms of their accuracy using the aforementioned 10-fold semi-cross-validation method. The accuracy was compared using the aforementioned semi-cross-validation method. In addition to the RMSE, the adjusted $R^2$ was used for accuracy. Figure 11 shows the distribution of the accuracy for each model and the results of multiple comparisons (Bonferroni, 0.05 level of significance). The accuracy of the FPV model that had a mean RMSE of 11,280 Yen and a mean adjusted $R^2$ of 0.79 was significantly higher than that of the other two models. However, the accuracy of the GF model was not significantly different from that of the baseline model.

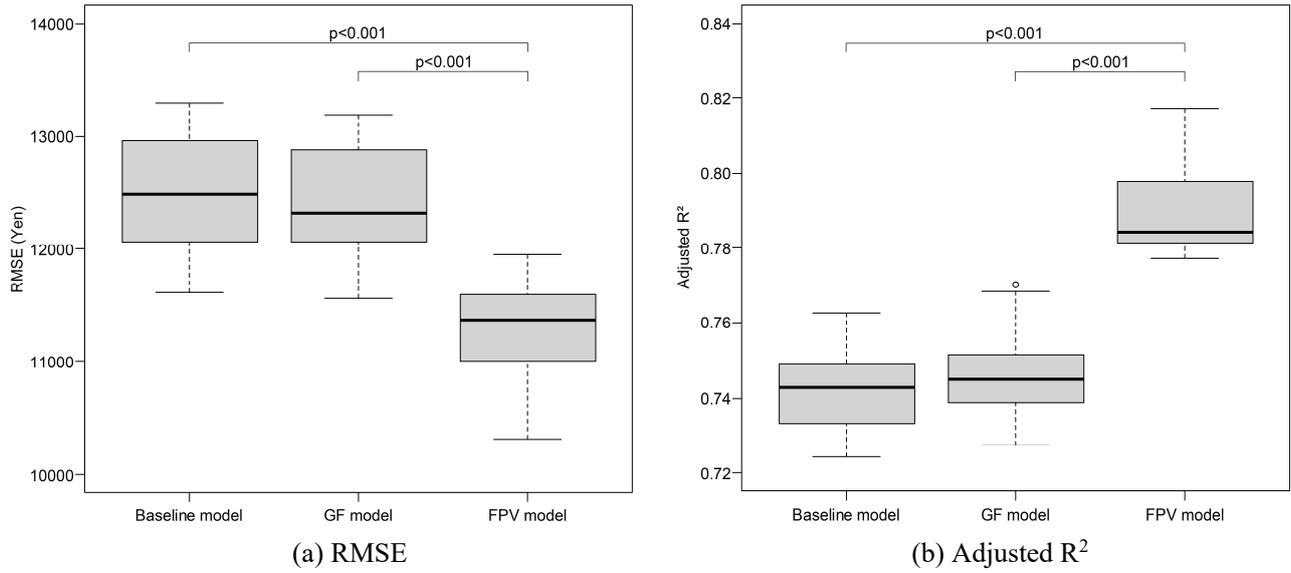

(a) RMSE  (b) Adjusted $R^2$

Figure 11. Comparison of the test data accuracy for 10-fold semi-cross-validations of each rent estimation model.

Tables 5 and 6 summarize the results of 10 and a half cross-validations on the test data of the model with FPV/graph attributes. In all models, except for the constant term, the most influential variables include surrounding land price, unit area, and year of completion, which is a general trend for Japanese real estate rent estimation models. In all models, the variance inflation factor (VIF) is less than 10, except for the constant term and categorical variable *Structure_RC*.

For the FPV model, the FPV has the next highest influence. As shown in Figure 12, the logarithmic values of the variables from #6 declined significantly. The most influential attribute in the GF model is *RRA_jp*, but its logarithmic value is as small as 1/20 of the FPV. However, it is easily interpretable; for example, a floor plan with a Japanese room that has a high degree of isolation has lower rent, or a floor plan with many closets has a higher rent.

Table 5. Statistics of the partial regression coefficients for 10-fold semi-cross-validations of the FPV model (values are expressed as mean ± SD).

| # | Variable | Coefficient | p | -log10(p) | VIF |
|---|---|---|---|---|---|
| 1 | *LandPrice* | 0.138 ± 0.005 | <0.000 ± 0.000 | 174.4 ± 22.16 | 1.260 ± 0.038 |
| 2 | *Area* | 1044 ± 50.83 | <0.000 ± 0.000 | 107.4 ± 6.085 | 1.201 ± 0.029 |
| 3 | *Year* | 929.8 ± 57.14 | <0.000 ± 0.000 | 79.74 ± 10.30 | 1.188 ± 0.037 |
| 4 | *Const* | −1.727E+06 ± 1.266E+05 | <0.000 ± 0.000 | 69.95 ± 10.69 | - |
| 5 | *FPV* | 599.0 ± 90.81 | <0.000 ± 0.000 | 68.29 ± 24.23 | 1.094 ± 0.036 |
| 6 | *L_Building* | 1073 ± 236.5 | <0.000 ± 0.000 | 18.81 ± 8.017 | 2.723 ± 0.312 |
| 7 | *Passenger* | 0.071 ± 0.011 | <0.000 ± 0.000 | 10.02 ± 2.834 | 1.038 ± 0.009 |
| 8 | *L_Room* | 838.3 ±368.1 | 0.001 ± 0.001 | 8.076 ± 5.397 | 2.064 ± 0.205 |
| 9 | *Distance* | −2.929 ± 0.963 | 0.025 ± 0.080 | 7.793 ± 3.018 | 1.085 ± 0.023 |
| 10 | *Structure_RC* | 3032 ± 2746 | 0.299 ± 0.189 | 0.634 ± 0.369 | 25.56 ± 5.765 |

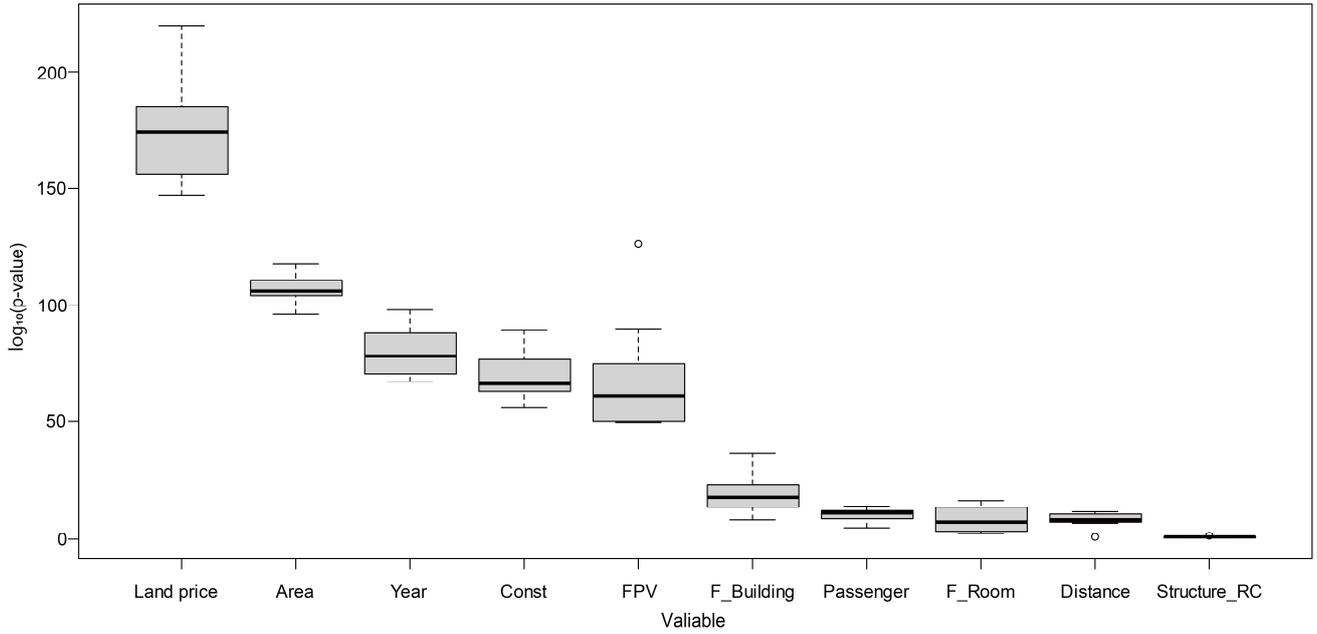

Figure 12. Distribution of the log values of each variable in the FPV model.

Table 6. Statistics of the partial regression coefficients for 10 semi-cross-validations of the FPV model (values are expressed as mean ± SD).

| # | Variable | Coefficient | p | −Log10(p) | VIF |
|---|---|---|---|---|---|
| 1 | *Land price* | 0.141 ± 0.007 | <0.000 ± 0.000 | 153.1 ± 19.40 | 1.280 ± 0.034 |
| 2 | *Area* | 1098 ± 86.27 | <0.000 ± 0.000 | 91.08 ± 9.045 | 1.337 ± 0.019 |
| 3 | *Const* | −1.869E+06 ± 1.691E+05 | <0.000 ± 0.000 | 56.81 ± 9.699 | - |
| 4 | *Year* | 926.6 ± 87.12 | <0.000 ± 0.000 | 55.74 ± 9.850 | 1.472 ± 0.051 |
| 5 | *F_Building* | 1299 ± 329.8 | <0.000 ± 0.000 | 22.73 ± 10.69 | 2.722 ± 0.304 |
| 6 | *Passenger* | 0.072 ± 0.013 | <0.000 ± 0.000 | 8.775 ± 2.973 | 1.046 ± 0.009 |
| 7 | *Distance* | −3.322 ± 1.100 | 0.019 ± 0.061 | 8.353 ± 3.493 | 1.087 ± 0.023 |
| 8 | *F_Dwelling* | 808.1 ± 471.1 | 0.058 ± 0.155 | 7.216 ± 5.988 | 2.072 ± 0.199 |
| 9 | *RRA_jp* | −3987 ± 1469 | 0.020 ± 0.038 | 3.100 ± 1.709 | 1.304 ± 0.053 |
| 10 | *Num_cl* | 1211 ± 614.0 | 0.102 ± 0.224 | 2.717 ± 2.034 | 3.230 ± 0.122 |
| 11 | *Num_bl* | 1533 ± 690.4 | 0.112 ± 0.168 | 1.634 ± 1.009 | 1.402 ± 0.052 |
| 12 | *RRA_wc* | 4082 ± 2343 | 0.195 ± 0.231 | 1.324 ± 1.026 | 2.174 ± 0.067 |
| 13 | *Num_edge* | −568.9 ± 250.1 | 0.148 ± 0.187 | 1.282 ± 0.828 | 3.503 ± 0.215 |
| 14 | *Depth_en* | 631.6 ± 1077 | 0.428 ± 0.324 | 0.737 ± 0.923 | 1.608 ± 0.042 |
| 15 | *RRA_bt* | −2354 ± 3229 | 0.450 ± 0.333 | 0.610 ± 0.657 | 2.082 ± 0.093 |
| 16 | *Structure_RC* | 2871 ± 3362 | 0.382 ± 0.259 | 0.595 ± 0.537 | 25.87 ± 5.706 |
| 17 | *H*$^*$ | −5471 ± 8045 | 0.593 ± 0.383 | 0.554 ± 0.870 | 2.212 ± 0.100 |

# Detailed analysis of the GCN model and its result

We obtained the FPVs of isomorphic access graphs using GCN at epoch 1550, which was learned using all 15,323 data points. We analyzed the relationship between the floor plans and the GCN model.

**Analysis 1: Differences in floor plans by FPVs**

The floor plans with the smallest, average, and largest FPVs are shown in Figure 13. The floor plans with the lowest FPV are unique because all living rooms are Japanese rooms, and all functions such as water supply are lined up in a row against the hallway. However, the average floor plan has features such as tatami mats facing a balcony and a separate kitchen, which provides the impression of being old-fashioned. The best floor plans are the most luxurious ones with two balconies, considerable storage space, and a high degree of spatial independence.

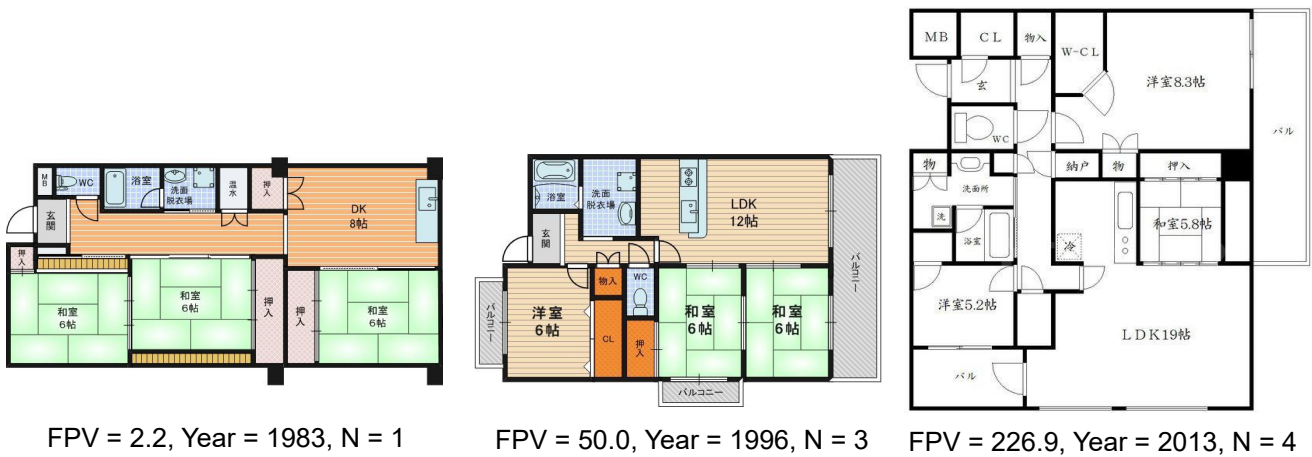

Figure 13. Examples of floor plans with low, mean, and high FPVs.

**Analysis 2: Basis for estimating the FPV model for each access graph**

The basis for estimating the FPV of the FPV model is determined. To this end, we visualized the access graphs of characteristic floor plans using IG, which is an explanatory method of deep learning models. Although there are several explanatory methods for deep learning models, IG was adopted because it has a clear theoretical foundation and is easy to implement.

First, the state in which there are no components in the access graph is set to 0, and the state in which the nodes and edges of the access graph to be examined are set to 1. The intensity of the graph is changed gradually from 0 to 1. IG expresses the contribution of each component of the access graph to the FPV as an integral value. The PyG graph model requires the input of nodes as a matrix of types, and the edges as a list of pairs of nodes at both ends of the matrix. The PyG graph model requires the input of nodes as a matrix of types, and the edges as a list of pairs of nodes at both ends of the matrix. The application of IG requires a simultaneous and continuous change in the strength of the components of the access graph. Captum adjusts the matrix such that continuous values of [0,1] are entered without any changes because the matrix of nodes indicates the use of the corresponding room by a binary value of {0,1}. Edges are represented by neighbor lists; therefore, their strengths as edges must be set separately. This was achieved by setting edge weights in the first convolutional layer of the graph. During the training and testing of the model, the weights of nodes and edges are fixed at 1, and their weights can be varied continuously in the range [0,1] by Captum only during the IG computation. The range of increments from 0 to 1 was set to 200 steps.

IG is applied to each unique access graph to determine the contribution of the components. The results are shown in Figure 14 for floor plans with low, average, and high FPVs visualized in a relatively easy-to-understand manner. Blue, gray, and red indicate a decrease, no change, and increase in the FPVs, respectively. The color scale is set for each property, and it is easy to understand that the connection of a Japanese room lowers the value of a floor plan with a low FPV. However, the result is similar for an average property. The difference is that balconies connected to Japanese rooms are rated lower for floor plans with low FPVs, whereas they are rated higher for floor plans with average FPVs. Nodes and edges around the dining room are evaluated positively in the floor plan with a high FPV, and this is consistent with the human evaluation of open-floor plans centered on the dining room. In contrast, the area around the water is rated as low. The dk_0 – la_0 – hw_0 loop seen in this floor plan is a looped flow line that has become popular in Japan in recent years. However, the FPV model does not recognize this loop as a good pattern, and there is a partial discrepancy between the loop and evaluations of the people.

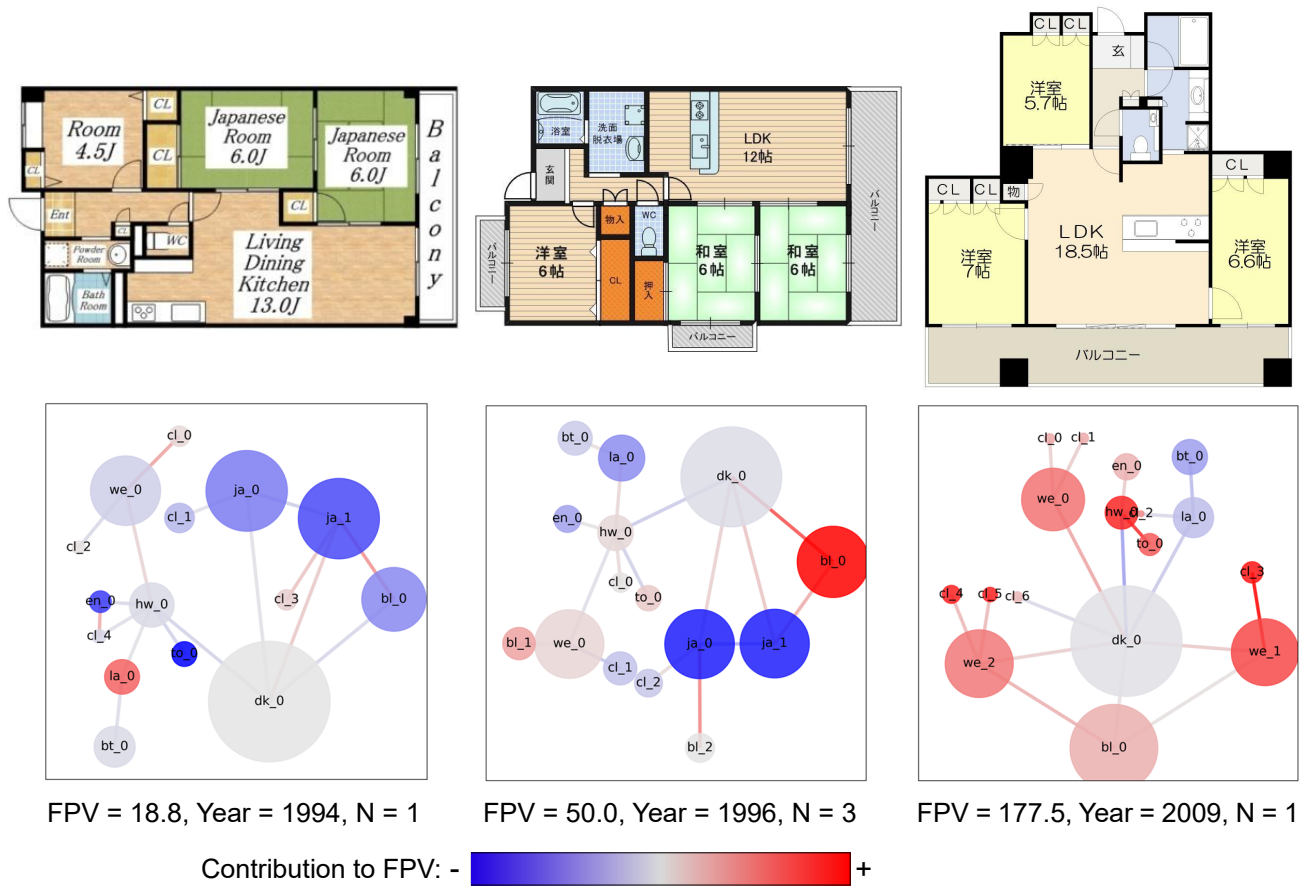

Figure 14. Visualization example of the contribution by the IG of access graphs for floor plans with low, medium, and high FPVs.

**Analysis 3: Understanding the overall estimation basis of the FPV model**

We characterized the estimation basis of the overall FPV model for all unique access graphs. IG is a relative value calculated for each floorplan, and it is not sensible to directly compare or aggregate the IG values of components with those of other floorplans. Therefore, the IG values of each floor plan component were standardized to have a mean of 0 and standard deviation of 1, and their distribution was tabulated for each component of all floor plans. The distributions were analyzed for significant differences from the overall mean at a significance level of 0.05 using the mean analysis method (ANOM; Nelson et al. 2005). The results are shown in Figure 15. Each point represents the mean value, and if the value is outside the range of the confidence interval [LDL, UDL], it is significantly different from the overall mean. The confidence interval decreases with an increase in the number of components.

The node with the largest positive contribution to the FPV is the balcony (0.463), whereas nodes with the largest negative contribution are the Japanese room (−0.267), dining kitchen (−0.216), and entrance (−0.165). The presence of a balcony may be interpreted as a guarantee of the minimum FPV because a balcony is an essential place to dry laundry in a family-oriented floor plan. As the popularity of Japanese rooms is declining in Japan, it is understandable that its presence has a negative impact. The dining room, kitchen, and entrance are common to all properties and are not involved in the differentiation factor if they are only present; therefore, it is difficult to interpret them as anything other than an overall adjustment factor.

For edges, the expression "bl-bt" for the edge type indicates that the edge connects the balcony and the bathroom. The edges with absolute values greater than 0.1 beyond the confidence interval are in the descending order from the highest positive contribution, closet-hallway (0.175), closet-dining kitchen (0.167), hallway-toilet (0.164), dining kitchen-Japanese room (0.152), and closet-entrance (0.121). The contribution tends to be higher when the closet is located at the hallway or entrance. The contribution is higher when the toilet is independent and entered from the hallway rather than that from the washroom. The Japanese room is entered from the dining room and kitchen rather than from the main hallway, which is the characteristic of modern layouts. Although these connections are convincing, the overall results suggest that the relationship between detailed accessory functions such as closets and toilets, instead of the relationship between rooms, tends to affect rent. The only significant edge with a negative contribution of −0.1 or less was the Japanese room-Japanese room (−0.138), which was less significant than the positive case. This result is easy to understand, although it suggests that the contribution of Japanese rooms tends to be lower when they are connected than when they stand alone, because of synergistic effects.

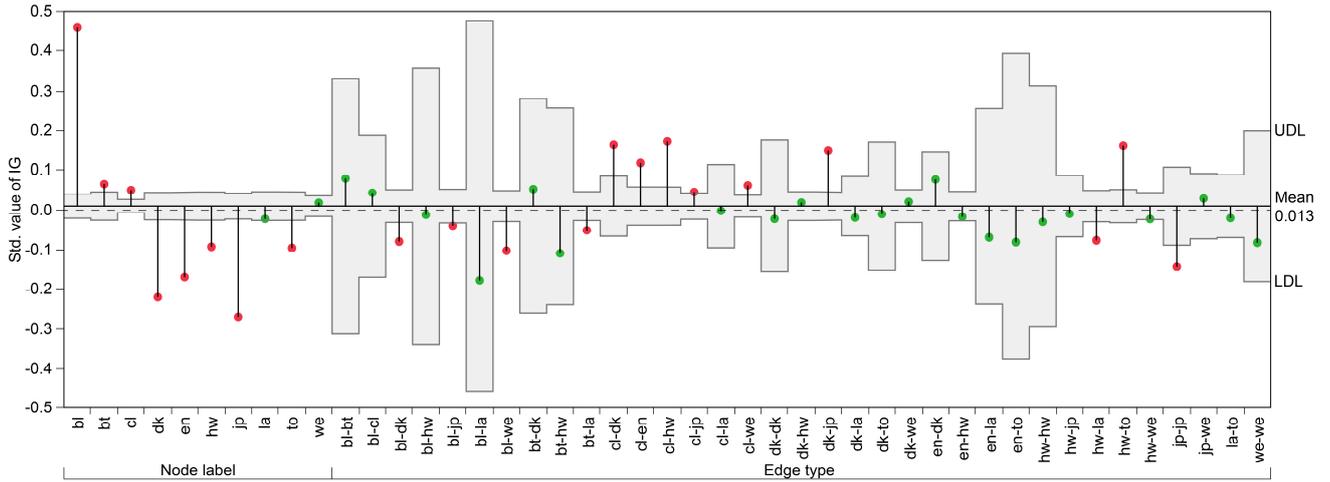

Figure 15. Distribution of node and edge contributions for all unique access graphs and confidence intervals for the mean analysis method.

## Discussion

Based on the proposed method and its results, we discuss the accuracy of the access graph data, relevance of the FPV model to human evaluations, and difference in use between the proposed method and an approach that makes the graph features explicit.

We discuss the accuracy of the access graph data. In this study, we improved the accuracy of access graph recognition by making detailed improvements to the methods used in previous studies; however, there are still cases of misclassification. In this method, the accuracy of semantic segmentation significantly affects the accuracy of neighboring graph extraction. Segmentation fails in small areas such as bathrooms and closets, when the text in the description is prominent and when the floor plan is black or white. The author personally heard that the accuracy of the semantic segmentation was improved in the newer version of the graph extraction method compared to the one we used; however, even so, perfect extraction may still be difficult to achieve. There are rare cases where the floor plan of a property is not written correctly in the first place. There are rare images of floor plans in which there are no doors or windows and the space is divided. Extracting access graphs from these images results in disconnected images, and continuous improvement in the quality of the access graph is required.

We explain the relationship between the FPV model and human evaluation. For example, the FPV and human evaluation may be similar in some respects and different in others, as seen in the aforementioned lack of detection of water circulation. The current GCN model does not appear to clearly detect the features of long flow lines beyond room pairs. However, this may be attributed to the visualization quirks of IG (Shrikumar et al. 2017), which is a general-purpose method such as IG but can output more intuitive results. GNNExplainer (Ying et al. 2019) is an explanatory method dedicated to GCNs. In this study, we used IG, which is theoretically superior and easy to implement; however, we believe that appropriate methods for explaining the GCN of floor plans need to be compared and investigated. The current access graphs represent only the connections between rooms with different uses. However, the presence or absence of openings was not considered except for the approach to the balcony. More basic features such as the size and shape of the rooms were not considered; it is highly possible that these points are the cause of the differences in human evaluations. The approach taken by previous studies, which addresses these issues using images rather than graphs that are difficult to convert, has the potential to improve the prediction accuracy, but the explanatory power of the model is limited. We believe that it should proceed within the framework of a graphical model to position it as a method for spatial analysis. This is a problem that needs to be improved from the graphing stage of floor plan images; however, we believe that it is an issue that needs to be addressed.

Finally, we discuss how to distinguish between explicit graph features and Space Syntax indices such as RRA, which are related to the centrality of space and are therefore appropriate for distinguishing lines of flow and public/private spaces. However, it is difficult to evaluate the relationship of connections between room uses. In addition, the approach to explicitly characterize the graph is not suitable for comprehensively evaluating the real estate value of a floor plan because the meaning of each indicator is easy to understand and suitable for grasping rough generalities. However, it is fragmentary and does not improve the explanatory power of the hedonic model, as shown in this study.

Therefore, it is not suitable for comprehensively evaluating the real estate value of floor plans. The use of a GCN is groundbreaking in that it can combine FPVs into a single value, which improves the explanatory power of the hedonic model. Deep learning models have higher-order nonlinearities, and even if the model explains the rationale, it does not always provide a generalized feature as graph features. Although this could be related to the aforementioned graph visualization, it is necessary to use different methods depending on whether the main objective is to evaluate the entire floor plan or specific features of the floor plan.

## Concluding remarks

We automatically extracted access graphs from a large number of floor plan images (over 40,000) for a family-oriented 3LDK rental apartment in Osaka Prefecture by improving existing extraction methods and obtained approximately 20,000 valid access graphs using a large Japanese real estate dataset. The, we performed isomorphism judgments on these access graphs and showed that there were nearly 8,000 unique access graphs. We also defined and implemented GCNs for the access graphs and estimated the FPV in terms of their contribution to rent. Furthermore, we compared the accuracy of the FPV model with a model that does not consider the access graph and a model that explicitly includes graph features in a semi-cross-validation test, and we found that the estimation accuracy of the FPV model was significantly better than those models, which yields a model with an adjusted $R^2$ of approximately 0.79. The features of floor plans that explain the rents from the learned GCNs were analyzed using the IG. We found that not only simple features such as the existence of balconies and Japanese rooms, but also the connection between closets and other rooms, independent toilets, and connectivity from the dining room to other rooms tend to increase the value of real estate, which can be learned inductively. The results showed that GCN can inductively learn that room connectivity, that is, connectivity from dining rooms to multiple rooms, tends to increase the value of real estate.

Although the modeling and analysis of floor plans using graphs have been conducted since the days of space syntax, these studies manually created graphs of floor plans, and only a relatively small number of graphs were analyzed. However, recent technological innovations centered on deep learning have made it possible to easily obtain a larger number of access graphs. Assuming the availability of such a large number of access graphs, this study proposed a method to model them using GCNs, estimated the real estate value of the floor plan, and showed that it is a powerful variable in the hedonic model. We showed that the GCN can provide a basis for estimation. We concluded that we proposed a new method for floor plan analysis and related fields, although some improvements are required.

## Appendix

A1. Basic method for extracting access graphs from images used in this study (Yamada et al. 2021) can be divided into two steps, as shown in Figure A1: (1) Semantic segmentation of rooms from a floor plan image and (2) a rule-based access graph extraction that recognizes the adjacency relationships among the segmented rooms through image processing.

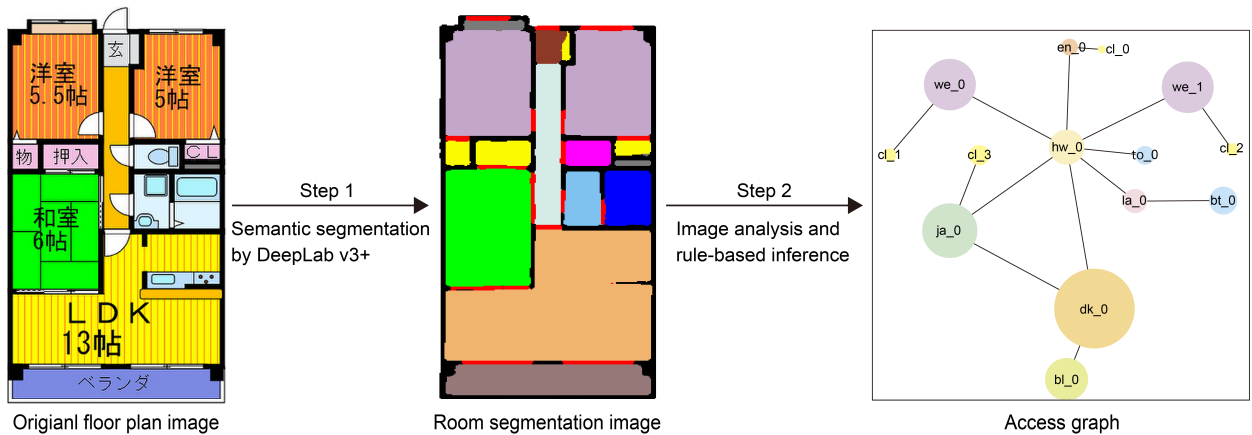

Figure A1. Basic procedure for extracting an access graph from a floor plan image.

However, this method often detects incorrect access graphs, and therefore, this study added the improvements shown in Table A1.

The extraction rate of the access graphs in Step 2 of Table A1 obtained by applying the modified extraction method was 19,998/46,466 = 0.43. A random sampling of 200 access graphs from the access graphs extracted in Step 2 was performed. The accuracy rate was determined by visual inspection to identify whether they were correct, as shown in Table A2. For comparison, the accuracy rate obtained using the original method was presented. The presence or absence of a closet was not considered an issue for the CL attached to the entrance because many of the drawings were too small to be considered as storage. Compared to the original extraction method, the improved access graph extraction method significantly improved the accuracy. However, the overall accuracy rate remained approximately $0.43 \times 0.81 = 0.35$.

Table A1. Improvements in access graph extraction methodology.

| Applicable steps | Description |
|---|---|
| Between Steps 1 and 2 | Since semantic segmentation blurs the outlines of walls, only black walls in the original image are extracted by image processing and superimposed on the segmented image. The extraction of walls from the original image is performed in the following order: gamma correction removal of the original floor plan image, grayscaling, binarization using Otsu's method, morphological transformation, and removal of small areas. |
| In Step 2 | In the segmentation image, if the boundaries by the doors in red are not straight but bend at right angles, the boundaries are divided into straight lines. |
| | Modified to not pull edges when rooms touching the door are not touching each other on both sides of the door. |
| | Edges between rooms that are unlikely to be connected as a general floor plan are deleted or modified, and the largest connected component of the modified graph is output as an access graph. |
| After completion of Step 2 | From the output access graph, a realistic access graph is extracted under the condition of a realistic number of nodes for a 3LDK floor plan in Japan. |

Table A2. Accuracy rate for 200 randomly sampled access graphs obtained from Procedure 2 in Table 1.

| Original | Modified |
|---|---|
| 0.42 | 0.81 |

A2. Let $a$, $b$, and $c$ represent the maximum RRA, mean RRA, and minimum $RRA$ of the access graph; $t = a + b + c$. The unrelative difference factor ($H$) was calculated as

$$H = -\frac{a}{t}\ln\left(\frac{a}{t}\right) + \frac{b}{t}\ln\left(\frac{b}{t}\right) + \frac{c}{t}\ln\left(\frac{c}{t}\right)$$

where $H^*$ normalizes the $H$ result into a scale between ln2 and ln3, and it is calculated as

$$H^* = -\frac{H - \ln(2)}{\ln(3) - \ln(2)}$$

# References

Adir A S, Berry J N and McGreal W S (1996) Hedonic modeling, housing submarkets and residential valuation. Journal of Property Research 13:67-83. DOI:10.1080/095999196368899.

Ahmed E H and Moustafa M N (2016) House price estimation from visual and textual features. In: Proceedings of the 8th International Joint Conference on Computational Intelligence 62-68. DOI:10.5220/0006040700620068.

Captum (2022) https://captum.ai/.

Chakrabarti D (2011) Graph Mining. In: Sammut, C., Webb, G.I. (eds) Encyclopedia of Machine Learning. Springer, Boston, MA. DOI:10.1007/978-0-387-30164-8_350.

Duvenaud D, Maclaurin D, Aguilera-Iparraguirre J, Gomez-Bombarelli R et al. (2015) Convolutional networks on graphs for learning molecular fingerprints. In: Proceedings of the 28th International Conference on Neural Information Processing Systems 2:2224-2232.

Bresson X and Laurent T (2017) Residual gated graph ConvNets. arXiv. DOI:10.48550/ARXIV.1711.07553.

Gao X, Asami Y, Zhou Y, Ishikawa T (2013) Preferences for floor plans of medium-sized apartments: A survey analysis in Beijing, China. Housing Studies, 28(3): 429-452. DOI:10.1080/02673037.2013.759542.

Glaeser E L, Kincaid M S and Naik N (2018) Computer vision and real estate: Do looks matter and do incentives determine looks. NBER Working Paper 27164. DOI:10.3386/w25174.

Goodman A C and Thibodeau T G (2007) The spatial proximity of metropolitan areas housing submarkets. Real Estate Economics 35:209-232. DOI:10.1111/j.1540-6229.2007.00188.x.

Hanazato T, Hirano Y and Sasaki M (2005) Syntactic analysis of large-size condominium units supplied in the Tokyo Metropolitan Area. Journal of Architecture and Planning (Transactions of AIJ), 591: 9-16. DOI: 10.3130/aija.70.9_5.

Hattori R, Okamoto K and Shibata A (2021) Impact analysis of floor-plan images for rent-prediction model. Journal of Japan Society for Fuzzy Theory and Intelligent Informatics 33(2): 640-650. DOI: 10.3156/jsoft.33.2_640.

Hillier B and Hanson J (1984) The Social Logic of Space. Cambridge, New York: Cambridge University Press. DOI:10.1017/CBO9780511597237.

Hofman E, Halman J I M and Ion R A (2006) Variation in housing design: Identifying customer preferences. Housing Studies 21:929-943. DOI:10.1080/02673030600917842.

Jim C Y and Chen W Y (2009) Value of scenic views: Hedonic assessment of private housing in Hong Kong. Landscape and Urban Planning 91(4):226-234. DOI: 10.1016/j.landurbplan.2009.01.009.

Jirovec R, Jirovec M M and Bosse R (1984) Architectural predictors of housing satisfaction among urban elderly men. Journal of Housing For the Elderly, 2(1): 21-32. DOI: 10.1300/j081v02n01_03 .

Kath N, Yamasaki T, Aizawa K and Ohama T (2020) Users' preference prediction of real estate properties based on floor plan analysis. IEICE Transactions on Information and Systems E103.D(2): 398-405. DOI: 10.1587/transinf.2019EDP7146.

Kipf T and Welling M (2016) Semi-supervised classification with graph convolutional networks. ArXiv. DOI: 0.48550/arXiv.1609.02907.

Kokusai Kogyo (2020) PAREA-Zip Kinki region.

Lovejoy K, Handy S and Mokhtarian P (2010) Neighborhood satisfaction in suburban versus traditional environments: An evaluation of contributing characteristics in eight California neighborhoods. Landscape and Urban Planning 97(1):37-48. DOI:10.1016/j.landurbplan.2010.04.010.

Marans R W and Spreckelmeyer K F (1982). Measuring overall architectural quality: A component of building evaluation. Environment and Behavior 14(6):652-670. DOI:10.1177/0013916582146002.

Michelson W M (1977) Environmental Choice, Human Behavior, and Residential Satisfaction. New York: Oxford University Press.

Ministry of Land, Infrastructure, Transport and Tourism (MLIT) (2015a) Official land price data. https://nlftp.mlit.go.jp/ksj/gml/datalist/KsjTmplt-L01-v2_3.html.

Ministry of Land, Infrastructure, Transport and Tourism (MLIT) (2015b) Prefectural land price survey data. https://nlftp.mlit.go.jp/ksj/gml/datalist/KsjTmplt-L02-v3_0.html.

Ministry of Land, Infrastructure, Transport and Tourism (MLIT) (2016) Data on the number of passengers by station. https://nlftp.mlit.go.jp/ksj/gml/datalist/KsjTmplt-S12-v2_3.html.

Narahara T and Yamasaki T (2022) Subjective functionality and comfort prediction for apartment floor plans and its application to intuitive online property search. IEEE Transactions on Multimedia. DOI: 10.1109/TMM.2022.3214072.

National Institute of Informatics (2015) LIFULL HOME'S Dataset, https://www.nii.ac.jp/dsc/idr/en/lifull/.

NetworkX developers (2014) https://networkx.org/.

Niepert M, Ahmed M, Kutzkov K (2016) Learning convolutional neural networks for graphs. In: Proceedings of the 33rd International Conference on International Conference on Machine Learning: 2014-2023.

Nelson P R, Wludyka P S, Copeland K A F (2005) The Analysis of Means. Society for Industrial and Applied Mathematics. DOI:10.1137/1.9780898718362.

Ostwald M J and Dawes M J(2018), The Mathematics of the Modernist Villa: Architectural Analysis Using Space Syntax and Isovists (Mathematics and the Built Environment, 3). Springer International Publishing AG. DOI: 10.1007/978-3-319-71647-3.

Poursaeed O, Matera T and Belongie S (2018) Vision-based real estate price estimation. Machine Vision and Applications 29: 667–676. DOI: 10.1007/s00138-018-0922-2.

PyG (2022) https://www.pyg.org/.

Rosen S (1974) Hedonic prices and implicit markets: Product differentiation in pure competition. Journal of Political Economy 82(1):34-55. DOI:10.1086/260169.

Simonyan K and Zisserman A (2015) Very deep convolutional networks for large-scale image recognition. In: 3rd International Conference on Learning Representations. DOI:10.48550/arXiv.1409.1556.

Solovev K and Pröllochs N (2021) Integrating floor plans into hedonic models for rent price appraisal. In: Proceedings of the Web Conference 2021, 2838-2847. DOI: 10.1145/3442381.3449967.

Sundararajan M, Taly A and Yan Q (2017) Axiomatic attribution for deep networks. In: Proceedings of the 34th International Conference on Machine Learning 70:3319–3328. DOI:10.5555/3305890.3306024.

Takizawa A, Yoshida K and Katoh N (2008) Applying graph mining to rent analysis considering room layouts, Journal of Environmental Engineering (Transactions of AIJ), 73(623): 139-146. DOI: 10.3130/aije.73.139.

Tamura J and Fang K (2022) Quality of public housing in Singapore: Spatial properties of dwellings and domestic lives. Architecture 2(1): 18-30. DOI: 10.3390/architecture2010002.

Yamada M, Wang X and Yamasaki T (2021) Graph Structure Extraction from Floor Plan Images and Its Application to Similar Property Retrieval. In: 2021 IEEE International Conference on Consumer Electronics (ICCE), Las Vegas, NV, USA: 1-5. DOI: 10.1109/ICCE50685.2021.9427580.

Yamasaki T, Zhang J and Takada T (2018) Apartment structure estimation using fully convolutional networks and graph model. In: Proceedings of the 2018 ACM Workshop on Multimedia for Real Estate Tech. DOI:10.1145/3210499.3210528.

Valente J, Wu S S, Alan G and Sirmans C F (2005) Apartment rent prediction using spatial modeling. Journal of Real

Estate Research 27(1). https://ssrn.com/abstract=953826.

Value Management Institute (2013) The Reality of the Rental Housing Market. https://www.mlit.go.jp/common/001011169.pdf.

You Q, Pang R, Cao L and Luo J (2017) Image-based appraisal of real estate properties. IEEE Transactions on Multimedia 19(12):2751-2759. DOI:10.1109/TMM.2017.2710804.

Zhao W, Xu C, Cui Z, Zhang T et al. (2018) When work matters: Transforming classical network structures to graph CNN. ArXiv, abs/1807.02653.

**Acknowledgments**

We thank Professor Toshihiko Yamazaki (Graduate School of Information Science and Technology, The University of Tokyo) for providing the code and trained models for extracting access graphs from the floor plan images obtained via the IDR Dataset Service of the National Institute of Informatics.

**Declaration of conflicting interests**

The author declared no potential conflicts of interest with respect to the research, authorship, and/or publication of this article.

**Funding**

This work was supported by a Grant-in-Aid for Scientific Research (grant number: 20K04872).

**ORCID iD**

Atsushi Takizawa: https://orcid.org/0000-0001-6474-6985

**Author biography**

Atsushi Takizawa, PhD., is Professor of Graduate School of Human Life and Ecology at Osaka Metropolitan University of Osaka, Japan. He has a PhD in Engineering from Kobe University. His research interests include spatial analytics in architecture and urban space, real estate analysis, evacuation planning, application of deep leaning, machine learning and discrete mathematics in architecture and urban planning. He teaches architectural planning, data science and computational design.